\documentclass[letterpaper, 10 pt, conference]{ieeeconf}  

\usepackage{amsmath}
\usepackage{amsfonts}
\usepackage{graphicx}
\usepackage{subcaption}
\usepackage{wrapfig}
\usepackage{booktabs}
\usepackage{multirow}
\usepackage{threeparttable}
\usepackage{float}
\usepackage{balance}
\usepackage{pgfplots}
\usepackage{hyperref}
\usepackage[listings]{tcolorbox}   
\pgfplotsset{compat=newest}
\usepackage{algorithm}
\usepackage{algpseudocode}

\IEEEoverridecommandlockouts                              
                                                          
\title{\fontsize{15.9}{20}\selectfont \bfseries AGRO-SUVIDE: Agentic Robotics for Surgical Viscoelastic Debridement}

\author{Shutong Jin$^{1,2}$, Ziyang Chen$^{1}$, Preethi Satish$^{1}$, Meadow Shen$^{1}$,\\ Gary Guthart$^{3}$, Florian T. Pokorny$^{2}$ and Ken Goldberg$^{1}$
\thanks{$^{1}$Department of Electrical Engineering and Computer Sciences, University of California, Berkeley, CA 94720, USA.}%
\thanks{$^{2}$Department of Robotics, Perception and Learning, KTH Royal Institute of Technology, Stockholm, SE 10044, Sweden.}%
\thanks{$^{3}$Intuitive Surgical, Sunnyvale, CA 94086, USA.}%
}

\begin{document}

\maketitle
\thispagestyle{empty}
\pagestyle{empty}

\begin{abstract}
Augmented dexterity has the potential to reduce the fatigue experienced by surgeons during repetitive surgical tasks. 
In this paper, we propose the first AGentic RObotics framework for SUrgical VIscoelastic DEbridement (AGRO-SUVIDE), the repeated removal of small fragments attached to a viscoelastic substrate.
Leveraging the self-improving and coding capability of agents, AGRO-SUVIDE adopts a modular framework.
Specifically, the demonstration analysis module automatically identifies recurring skills from a single expert demonstration, using both visual and kinematic information.
The construction module then builds each skill, either as a procedural model-based skill the agent codes against a scaffolded library or as a model-free policy-based skill. 
At runtime, the monitoring module composes the skills into a loop-style graph sized to the number of fragments it observes, then verifies pre- and post-conditions of each skill to decide whether to advance or retry.
We evaluate AGRO-SUVIDE through 340 physical trials on the da Vinci Research Kit (dVRK). 
AGRO-SUVIDE achieves an average single-fragment removal success rate of 85\%, completing consecutive three-fragment removal at 60\% and at 95\% with one human intervention.
It further generalizes to unseen five-fragment scenarios with an average success rate of 80\% for single-fragment removal.
Project page: \href{https://surgical-robotics.github.io/AGRO-SUVIDE/}{https://surgical-robotics.github.io/AGRO-SUVIDE/}
\end{abstract}

\section{INTRODUCTION}
Surgical robots such as the da Vinci system (Intuitive Surgical, Inc., USA) are now widely deployed, offering enhanced vision and improved precision in minimally invasive procedures~\cite{yip2023artificial}.
These systems remain largely teleoperated, with outcome quality dependent on the individual surgeon, and fatigue may accumulate over long procedures composed of repetitive subtasks.
Augmented Dexterity, proposed by Goldberg and Guthart~\cite{goldberg2024augmented}, envisions the robot executing surgical subtasks under close surgeon supervision, reducing both performance variability across surgeons and the fatigue of repetitive work.

Surgical debridement, the removal of infected or dead tissue fragments from a wound, is a fundamental yet highly repetitive procedure in wound care and burn treatment~\cite{amadeh2025comparative}.
It typically follows a fixed loop in which the surgeon lifts each fragment, cuts it free, and clears it into a receptacle before moving to the next.
Repetition of this kind may cause fatigue over a long procedure, while a single overlooked fragment can lead to dangerous infection.
Its procedural regularity makes debridement a natural application for augmented dexterity, and prior work has explored automation of this loop~\cite{kehoe2014autonomous,murali2015learning,chen2026macaw}.
Prominent examples encode the loop as a finite state machine, in which a human manually segments a demonstration into motion skills and defines the conditions that advance execution between them~\cite{kehoe2014autonomous,murali2015learning}.
However, covering the range of conditions encountered requires repeated rounds of tuning on the robot, where failures are replayed, localized, and often corrected by hand, leaving cost and quality tied to individual expertise.
Imitation learning offers an alternative, replacing hand-crafted logic with policies trained directly from demonstration, and has been applied to surgical subtasks including needle picking and tissue lifting~\cite{haworth2026suturebot}. 
However, learned policies accumulate error over long horizons~\cite{ross2011reduction, belkhale2023data}, and consecutive debridement of multiple fragments compounds small deviations with each removal. 
Training such policies also requires demonstration data at a scale that is costly to collect on a physical system.

\begin{figure}[t]
    \centering
    \includegraphics[width=\linewidth]{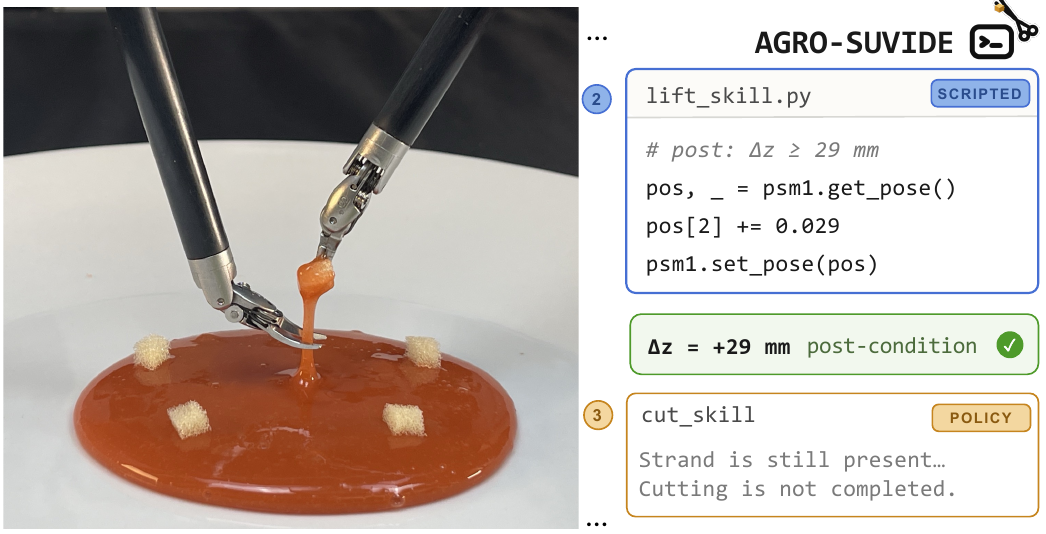}
    \caption{\small AGRO-SUVIDE: Agents automatically identify and improve modular skills from one expert demonstration, build each as model-based or model-free, and a monitor executes them at runtime.
    Here the monitor confirms that the model-based skill ``lift" has met its post-condition of $\Delta z \geq 0.029$ m, and holds the model-free skill ``cut" running until its post-condition is met. }
    \label{fig:first_figure}
    \vspace{-1em}
\end{figure}

Recently, agentic systems have advanced rapidly on the strength of self-improvement, with results spanning repository-scale software engineering~\cite{yang2024swe} and scientific discovery~\cite{novikov2025alphaevolve}.
This capability results from a loop-style process in which the agent writes code, executes it, reads the resulting artifacts, and revises the implementation, with flexible tool use at every stage. 
Robotics has begun to develop agentic systems with self-improving capabilities, supplying the artifact feedback that the loop requires either through simulation or through resetting infrastructure on physical robots~\cite{chen2026gap, lu2026aspire, xiao2026enpire, fu2026cap}.
In this work, we ask: 

\textit{Can the self-improving capability of agents exploit the repetition inherent in debridement to identify and improve modular skills automatically? And how do we ensure the identified skills are executable and chain reliably?}

Inspired by this, we propose AGRO-SUVIDE, the first agentic robotics framework for surgical augmented dexterity (\textit{Fig.}~\ref{fig:first_figure}). 
Exploiting the repetition inherent in debridement, a demonstration analysis module analyzes one expert three-fragment demonstration to identify and improve modular skills.
Each identified skill is bounded by explicit pre- and post-conditions defined over vision and kinematics.
The construction module reads those conditions and builds each skill to satisfy them, either as a model-based skill coded against a scaffolded library exposing the dVRK API alongside foundation models such as SAM3~\cite{carion2026sam}, or as a model-free policy-based skill where a model-based one cannot satisfy the conditions reliably.
At runtime, the monitoring module composes the skills into a loop-style graph sized to the number of fragments it observes, then verifies the pre- and post-conditions of each skill to decide whether to advance or retry. 
Our contributions are fourfold:
\begin{itemize}
    \item We propose AGRO-SUVIDE, the first agentic robotics framework for surgical viscoelastic debridement, executing the task through a combination of model-based and model-free skills.
    \item We propose a demonstration analysis pipeline that automatically identifies modular skills from a single demonstration. It exploits the repetition inherent in surgical debridement to identify and improve each skill's pre- and post-conditions over vision and kinematics.
    \item We propose a compositional data collection protocol that collects bimanual demonstrations as per-skill single-arm segments, allowing model-based and model-free skills to chain without adjustment.
    \item We evaluate AGRO-SUVIDE over 340 physical trials, achieving an 85\% single-fragment success rate and a 60\% consecutive three-fragment success rate, and generalizing to unseen five-fragment scenarios at 80\%.
\end{itemize}

\section{RELATED WORK}
\subsection{Surgical Augmented Dexterity } 
Goldberg and Guthart~\cite{goldberg2024augmented} propose augmented dexterity as a palatable alternative to ``supervised autonomy" in robotic surgery. Surgical augmented dexterity has been approached predominantly through a modular pipeline of perception, planning, and control with explicitly specified transitions~\cite{attanasio2021autonomy}.
One example is Murali et al.~\cite{murali2015learning}, who demonstrate multilateral cutting and debridement by segmenting an expert demonstration into primitives, encoding them as a finite state machine on the dVRK, and advancing between states on manually set values such as a color threshold for fragment detection. 
Other typical surgical subtasks that leverage the same augmented dexterity pipeline include suturing~\cite{hari2025stitch}, tumor resection~\cite{pore2021learning}, and knot tying~\cite{chen2025surgical}. 
Such implementations execute quickly, run deterministically, and remain fully inspectable, but they rely heavily on the expertise of the engineer who tunes them.
A complementary line of work replaces hand-specified logic with policies learned directly from demonstration.
SRT~\cite{kim2024surgical} shows that a shared imitation learning recipe can be applied across several surgical subtasks on the dVRK, and subsequent work has extended this to retraction, needle handling, and end-to-end suturing~\cite{haworth2026suturebot, moghani2025sufia}.
Recent hierarchical frameworks have noted that errors accumulate over task execution time and therefore propose separate planning from execution to better handle long-horizon tasks. 
SuFIA~\cite{moghani2024sufia} pairs LLM reasoning with perception modules, and SRT-H~\cite{kim2025srt} learns policies at both levels, with a high-level policy for task planning and a low-level policy for robot execution. 
Both systems fix a single implementation class across the library, model-based in SuFIA and model-free in SRT-H, leaving the conditions between skills either hand-specified or absorbed into a learned planner.
AGRO-SUVIDE instead leverages the inherent repetition in surgical debridement to automatically identify the skills from a single expert demonstration, each with explicit pre- and post-conditions over vision and kinematics. 
Model-based and model-free skills then coexist in one loop-style execution graph.

\subsection{Agentic Robotics Systems}

Agentic coding for robotics \cite{liang2023code} offers a promising paradigm for integrating model-based and model-free approaches to enable more interpretable and reliable robot task execution \cite{chen2026gap}. Code as Policy (CAP) \cite{liang2023code} pioneers the use of agentic coding to compose high-level motion skills and generate new policy code for robot manipulation. As an extension, CAP-X \cite{fu2026cap} introduces a unified framework for systematically studying CAP agents across simulated and real-world robotic tasks. Graph-as-Policy~\cite{chen2026gap} proposes an agentic harness that generates executable graphs from a reusable skill library and employs internal simulation rehearsal to optimize graph composition, improving success rates and throughput in robotic manipulation tasks. Similarly, Xu et al. \cite{xu2026reaction} introduce a multi-agent framework for generating directed task graphs with failure recovery capabilities. ASPIRE \cite{lu2026aspire} enables continual learning through the autonomous discovery of reusable skills across diverse simulated tasks, while ENPIRE \cite{xiao2026enpire} focuses on the self-improvement of robot policies in the real world.
AGRO-SUVIDE applies this self-improving capability to a new setting, using the repetition inherent in surgical procedures as the supervision signal that lets an agent identify recurring skills and turn them into executable implementations.

\subsection{Surgical Phase Recognition}
Online surgical phase recognition provides real-time semantic understanding of the surgical procedure, enabling context-aware technologies such as autonomous surgical manipulation, image-guided navigation and intraoperative decision-making \cite{chen2024towards}. Two main data modalities are commonly exploited for surgical phase recognition: vision and kinematics \cite{hutchinson2023evaluating}. Vision-based approaches \cite{hu2024ophnet} have been widely explored, as surgical images and video clips provide rich semantic cues about the ongoing procedure. Jin et al. \cite{jin2021temporal} designed a temporal memory relation network with a long-range memory bank to recognize surgical phase from videos. Funke et al. \cite{funke2025tunes} introduced a temporal U-Net with self-attention to recognize surgical phase from video clips, achieving state-of-the-art recognition accuracy in cholecystectomy. In contrast, Nicolo et al. \cite{pasini2023grace} leveraged kinematics data from the two da Vinci robot arms and developed a LSTM-based network to recognize surgical phase for autonomous endoscope navigation. Experiments demonstrated that their approach could relieve the operator from manually manipulating the endoscope while performing a suturing task. Some studies \cite{van2021gesture} have also explored multimodal approaches that combine visual and robot kinematic information for surgical phase recognition. Bai et al. \cite{bai2025multimodal} designed a multimodal disentanglement graph network to encode heterogeneous features and a calibrated prediction decoder to recognize surgical phase, demonstrating the effectiveness of multimodal information for surgical phase recognition. However, prior neural network-based approaches rely heavily on human effort to annotate specific surgical phases for model training, making the annotation process labor-intensive and time-consuming. AGRO-SUVIDE explores the possibility of using agents to identify phase recognition skills from multimodal data, including vision and kinematics, without requiring human-provided labels.

\section{PROBLEM FORMULATION}
\subsection{Task Description}
We study surgical debridement, the removal of infected or dead tissue fragments from a wound. 
We keep the task close to the clinical setting by placing each fragment in a viscoelastic substrate.
Adhesive strands continue to bind the fragment when it is lifted, so removal typically requires severing rather than lifting alone.
We perform the task on the dVRK, using its two Patient-Side Manipulators (PSMs). 
PSM1 (right) is mounted with an instrument of Large Needle Driver gripper, while PSM2 (left) carries Curved Scissors.
Each fragment is cleared by the same sequence.
PSM1 grasps the fragment and lifts it clear of the substrate.
PSM2 cuts the strand holding it to the substrate. 
PSM1 then carries the fragment to the receptacle and wipes it in together with the residual strand, before returning for the next fragment.
This repeats until no fragment remains embedded in the substrate.

\subsection{Problem Definition}
We adopt the following notation:
\begin{itemize}
\item $\mathcal{S}$: the viscoelastic substrate. 
\item $\mathcal{F}$: the set of fragments in the substrate. Each fragment $f \in \mathcal{F}$ is made of foam rubber rather than viscoelastic material, so it does not break apart and is removed as a unit, keeping $\mathcal{F}$ discrete. Fragments are partially embedded in the viscoelastic substrate. When a fragment is grasped and lifted, some viscoelastic material remains attached to the fragment and requires cutting with the surgical scissors to free the fragment. 
\item $\mathcal{R}$: the receptacle for removed fragments.
\item $\mathcal{V}$: the visual stream, a sequence of stereo frame pairs $\mathcal{V} = \{v_t\}_{t=1}^{T}$ with $v_t = (v^{\text{L}}_t, v^{\text{R}}_t)$, where $v^{\text{L}}_t$ and $v^{\text{R}}_t$ are the left and right images captured at timestep $t$.
\item $\mathcal{P}$: the kinematic stream, a sequence of states $\mathcal{P} = \{p_t\}_{t=1}^{T}$ with $p_t = (p^{1}_t, p^{2}_t)$, where $p^{i}_t = (j^{i}_t, g^{i}_t, x^{i}_t)$ gives the joint angles $j^{i}_t \in \mathbb{R}^{6}$, jaw angle $g^{i}_t \in \mathbb{R}$, and end-effector pose $x^{i}_t \in \mathrm{SE}(3)$ of PSM$i$ at timestep $t$, with $x^{i}_t$ obtained from $j^{i}_t$ by forward kinematics.

\item $\pi$: the policy. A single-arm policy maps $(v^{\text{L}}_t, j^{i}_t, g^{i}_t)$ to the joint and jaw commands $(\hat{j}^{i}_t, \hat{g}^{i}_t)$ for arm $i$; a bimanual policy maps $(v^{\text{L}}_t, j^{1}_t, g^{1}_t, j^{2}_t, g^{2}_t)$ to $(\hat{j}^{1}_t, \hat{g}^{1}_t, \hat{j}^{2}_t, \hat{g}^{2}_t)$. 

\item $\mathcal{E}$: the expert demonstration, comprising synchronized streams $\{v^{\text{L}}_t\}_{t=1}^{T_\mathcal{E}}$ and $\{p_t\}_{t=1}^{T_\mathcal{E}}$ recorded over one execution of length $T_\mathcal{E}$ and containing $N$ fragment removals.

\item $\mathcal{B}$: the scaffolded library available to the agent containing dVRK control APIs and foundation models.

\item $\mathcal{T}$: the artifacts, $\mathcal{T} = \{\mathcal{T}_s\}$, where $\mathcal{T}_s$ records what stage $s$ produced and which checks passed. Retaining these localizes a failed constraint to the stage responsible.

\item $\mathcal{K}$: the skills, where each $\kappa \in \mathcal{K}$ is a unit of motion on one arm, $\kappa = (\texttt{PRE}, \texttt{POST}, \texttt{DESC})$, with $\texttt{PRE}$ and $\texttt{POST}$ its pre- and post-conditions over vision and kinematics, and $\texttt{DESC}$ the motion between them.

\item $\Phi$: the phases, where each $\phi \in \Phi$ pairs a skill on PSM1 with one on PSM2, $\phi = (\kappa^{1}, \kappa^{2})$ with $\kappa^{i} \in \mathcal{K}$.

\item $\mathcal{L}$: the loop-style graph, whose body is an ordered sequence of phases $\phi \in \Phi$ removing a single fragment, repeated once for each of the $n$ fragments observed at runtime. $n$ is independent of $N$.
\end{itemize}

Given a single expert demonstration $\mathcal{E}$, the demonstration analysis module identifies the skills $\mathcal{K}$ together with the conditions ($\texttt{PRE}$, $\texttt{POST}$, $\texttt{DESC}$) of each $\kappa \in \mathcal{K}$. 
The construction module then builds each $\kappa$ against these conditions, either as a procedural model-based skill coded against the scaffolded library $\mathcal{B}$ or as a model-free policy-based skill with a trained policy $\pi$. 
At runtime, the monitoring module pairs skills into phases $\Phi$ and composes the loop-style graph $\mathcal{L}$ sized to the $n$ fragments observed, then verifies $\texttt{PRE}$ and $\texttt{POST}$ of each skill to decide whether to advance or retry.
Execution of $\mathcal{L}$ completes when no fragment $f \in \mathcal{F}$ remains in $\mathcal{S}$.

\begin{figure*}[t]
    \centering
    \includegraphics[width=\linewidth]{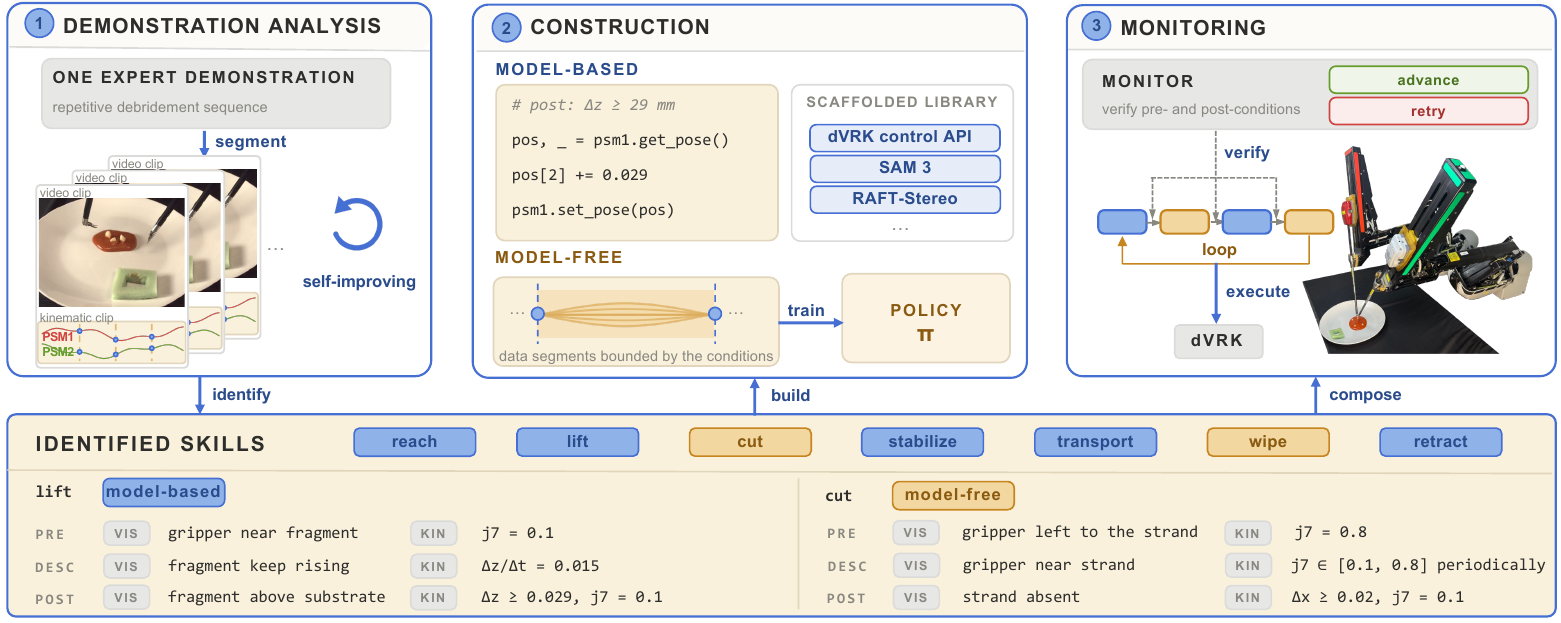}
    \caption{\small AGRO-SUVIDE adopts a modular design. Given one expert demonstration, the demonstration analysis module exploits the repetition in surgical debridement to identify and improve modular skills with explicit pre- and post-conditions defined over vision and kinematics. The construction module then reads those conditions and builds each skill to satisfy them, either as a procedural model-based skill coded against a scaffolded library or as a model-free policy-based skill trained on data segments bounded by the same conditions. At runtime, the monitoring module composes the loop-style graph sized to the number of fragments it observes, and verifies the conditions to decide whether to advance or retry.}
    \vspace{-1em}
    \label{fig:structure}
\end{figure*}

\subsection{Assumptions}
We make the following assumptions:
\begin{itemize}
    \item The expert demonstration $\mathcal{E}$ contains no failure recovery, such as reaching toward the wrong fragment and correcting mid-motion, and no idiosyncratic execution, such as pausing in the reaching trajectory.
    \item The expert demonstration $\mathcal{E}$ is procedurally repetitive: each of its $N$ fragments is removed by the same ordered sequences, consistent with clinical practice, rather than by sequences differing in order or composition.
    \item The structural components of the scene remain unchanged between demonstration and evaluation, including receptacle placement and camera pose. Variation in the shape and size of substrate $\mathcal{S}$, and in the pose and size of each fragment $f \in \mathcal{F}$, is
    permitted.
    \item Every fragment $f \in \mathcal{F}$ is at least partially exposed on the surface of the substrate $\mathcal{S}$, so that it is visible to the perception models.
\end{itemize}

\section{METHODOLOGY}
This section describes the three modules that constitute AGRO-SUVIDE, illustrated in \textit{Fig.}~\ref{fig:structure}.

\subsection{Demonstration Analysis Module}
We leverage the repetition inherent in surgical debridement as the supervision signal to automatically identify and improve modular skills.
The agents explore under two constraints: (a) the boundary constraint, requiring that all skills assigned the same semantic label share the same pre- and post-conditions over visual and kinematic states; 
(b) the repetition constraint, requiring that each $\phi \in \Phi$ recur $N$ times and in the same order within $\mathcal{E}$.
To bound the search, the demonstration analysis pipeline is structured into four stages, with each stage emitting an artifact $\mathcal{T}_s$ that the agent revises against when either constraint fails.

\paragraph{Stage 1: Pre-Processing} 
From the expert demonstration $\mathcal{E}$, the agent analyzes visual stream $\{v^{\text{L}}_t\}_{t=1}^{T_\mathcal{E}}$ jointly with each arm's kinematic steam, forming the pairs $\big(\{v^{\text{L}}_t\}_{t=1}^{T_\mathcal{E}}, \{p^{1}_t\}_{t=1}^{T_\mathcal{E}}\big)$ and $\big(\{v^{\text{L}}_t\}_{t=1}^{T_\mathcal{E}}, \{p^{2}_t\}_{t=1}^{T_\mathcal{E}}\big)$.
Within each pair, it detects salient changes, defined as transitions in the spatial relationship between scene components or in the kinematic state, such as the gripper entering the receptacle or the jaw beginning to open. 
A change in either stream marks a boundary, and the clips between consecutive boundaries form the raw skills. 
This yields two sets $\mathcal{K}^{1}_\text{raw}$ and $\mathcal{K}^{2}_\text{raw}$, one per arm, with the criterion adopted for each recorded in $\mathcal{T}_1$.

\paragraph{Stage 2: Semantic Merge} 
For each raw skill in $\mathcal{K}_\text{raw}$, the agent queries a vision-language model from $\mathcal{B}$ over the frames it spans and receives a natural-language description of what occurs within it, such as ``a fragment is gripped and lifted upward, revealing the strand still binding it to the gel". 
A second agent compares the returned descriptions, merges raw skills whose descriptions are similar, and assigns each merged set a canonical name such as ``lift". 
The raw skills sharing one name form a candidate skill $\kappa \in \mathcal{K}_\text{candidate}$.
Criteria and uncertainties raised during description or merging are documented in $\mathcal{T}_2$.

\paragraph{Stage 3: Improving under the Boundary Constraint} 
For candidate skills in $\mathcal{K}_\text{candidate}$, the agent composes
\texttt{PRE} and \texttt{POST} separately on the visual and kinematic channels, pairing a visual condition with kinematic one, such as the jaw in contact with the fragment together with a jaw value of 0.1. 
\texttt{DESC} describes what occurs between them, covering jaw and arm state and velocity.
The agent reports the fraction of instances satisfying \texttt{PRE} and \texttt{POST} and flags those that do not. Flagged instances are returned for improvement, where the visual relations and numeric kinematic conditions are tightened until all instances agree.
The agreement rate at each iteration, together with the conditions that produced it, is documented in $\mathcal{T}_3$.

\paragraph{Stage 4: Improving under the Repetition Constraint} 
The demonstration contains the same procedure repeated $N$ times, so the agent checks that the identified phase sequence $\phi \in \Phi$  likewise appears $N$ times. Where it does not, the disagreement localizes to specific raw skills, which return to Stage 2 for re-proposal and pass through Stage 3 again.
The recovered sequence and the raw skills returned for re-proposal are documented in $\mathcal{T}_4$.

\paragraph{Identified Skills} 
Once both constraints are met, the agent emits the skill set $\mathcal{K}$, where each $\kappa \in \mathcal{K}$ is $(\texttt{PHASE}, \texttt{PRE}, \texttt{POST}, \texttt{DESC}, \texttt{PARAM}, \texttt{TMOUT})$.
\texttt{PHASE} names the phase the skill belongs to.
\texttt{PRE} and \texttt{POST} are dictionaries rather than single predicates: each holds a set of key-value clauses over the visual and kinematic channels, and a skill typically carries several in each. 
\texttt{PARAM} gives a nominal value and a tunable range per parameter, taken as the median and the padded spread over skills in the same group. 
\texttt{TMOUT} derives a deadline from the longest observed duration with a
margin, so a slower-than-median execution is tolerated rather than cut off.
The same conditions are consumed by the authoring and monitoring modules; the identified skills and two identified examples appear in \textit{Fig.}~\ref{fig:structure}, with one condition per channel for brevity.

All four stages are specified to the agent in a textual specification $\mathcal{M}$, written in Anthropic's \texttt{SKILL.md}~\cite{anthropic2025agentskills} format. \textit{Alg.}~\ref{alg:discovery} gives an illustration of the demonstration analysis pipeline.

\begin{tcolorbox}[
  colback=gray!6, colframe=gray!25, boxrule=0.4pt, arc=2pt,
  left=6pt, right=4pt, top=5pt, bottom=5pt,
  boxsep=0pt, width=\linewidth,
  fontupper=\scriptsize]
\begin{verbatim}
contract <skill> : <arm>
  PHASE  <phase name>
  PRE    vis { <condition>, ... }
         kin { <condition>, ... }
  DESC   vis { <region relations over the skill> }
         kin { <jaw/arm state and velocity> }
  POST   vis { <condition>, ... }
         kin { <condition>, ... }
  PARAM  <name> = <nominal> <unit> in [<lo>,<hi>]
  TMOUT  <factor> x nominal duration
\end{verbatim}
\end{tcolorbox}



\begin{algorithm}[h]
\caption{Demonstration Analysis}
\label{alg:discovery}
\footnotesize
\begin{algorithmic}[1]
\Require expert demonstration $\mathcal{E}$, repetition count $N$, textual specification $\mathcal{M}$
\State $\mathcal{K}_\text{raw}, \mathcal{T}_1 \gets \textsc{Pre-Process}(\mathcal{E})$
\State \hfill \textit{// Stage 1: Pre-Processing}
\Repeat
  \State $\mathcal{K}_\text{candidate}, \mathcal{T}_2 \gets \textsc{Merge}\big(\textsc{Label}(\kappa_\text{raw})\
         \forall \kappa_\text{raw} \in \mathcal{K}_\text{raw}\big)$
 \State \hfill \textit{// Stage 2: Semantic Merge}
  \Repeat
    \State $\mathcal{K}_\text{candidate}, \mathcal{T}_3 \gets \textsc{Tighten}(\mathcal{K}_\text{candidate})$
    \State \hfill \textit{// Stage 3: Boundary Constraint}
  \Until{all instances of each $\kappa_\text{candidate} \in \mathcal{K}_\text{candidate}$ satisfy (\texttt{PRE}, \texttt{POST})}
  \State $\Phi, \mathcal{T}_4 \gets \textsc{Phases}(\mathcal{K})$
\State \hfill \textit{// Stage 4: Repetition Constraint}
\Until{$\phi \in \Phi$ recurs $N$ times}
\State \Return $\mathcal{K} \gets \mathcal{K}_\text{candidate}$
\end{algorithmic}
\end{algorithm}

\subsection{Construction Module}
The construction module implements each skill against its conditions, with \texttt{PRE} and \texttt{POST} bounding it and \texttt{DESC} specifying the motion between.
Each skill is implemented in one of two forms: a procedural model-based skill or a model-free policy-based skill.

\paragraph{Procedural Model-Based Skills}
Procedural model-based skills are coded directly by the agent over a scaffolded library $\mathcal{B}$. 
The library exposes the dVRK API for bimanual control, with calls to set Cartesian end-effector targets, joint and jaw velocity ratios, jaw state for either arm, etc. It also exposes foundation models for perception, such as SAM3 for semantic segmentation~\cite{carion2026sam}, and RAFT-Stereo for depth estimation~\cite{lipson2021raft}.
Each entry carries a description of its inputs, outputs, and function, so the agent can select among them.

\begin{figure}[!t]
    \centering
  \begin{subfigure}[h]{0.48\textwidth}
    \centering
    \includegraphics[width=\textwidth]{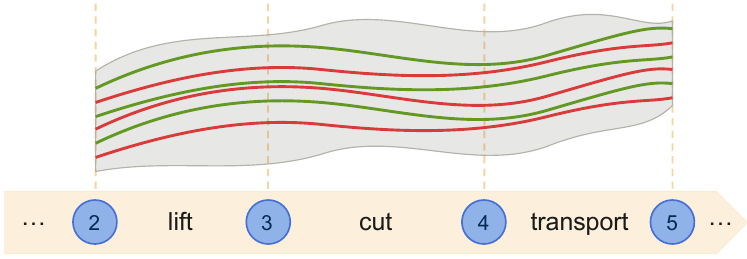}
    \vspace{-0.5 em}
    \caption{}
    \label{fig:dataA}
  \end{subfigure}
  \begin{subfigure}[h]{0.48\textwidth}
    \centering
    \includegraphics[width=\textwidth]{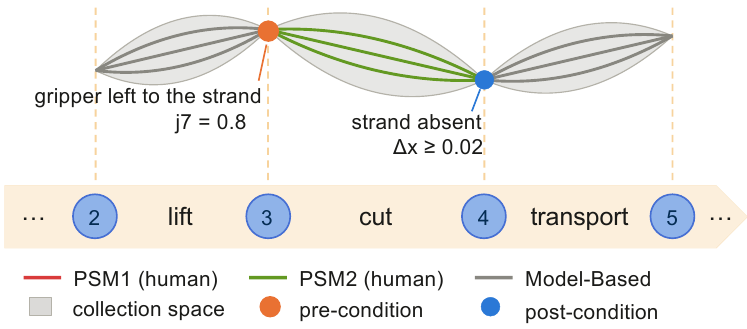} 
      \vspace{-0.5em}
    \caption{}
    \label{fig:dataB}
  \end{subfigure}
    \caption{\small Two data collection protocols: (a) conventional collection, where the operator teleoperates both arms continuously with no boundary between skills; (b) the proposed compositional collection, where the agent drives one arm to a state satisfying \texttt{PRE} (gripper left to the strand, $j = 0.8$ m), the operator demonstrates a single skill under \texttt{DESC}, and the agent verifies \texttt{POST} (strand absent, $\Delta x \geq 0.02$ m). Only segments required for policy-based skills are recorded.}
    \label{fig:data_collection}
    \vspace{-1em}
\end{figure}

\begin{figure*}[!t]
    \centering
    \includegraphics[width=\linewidth]{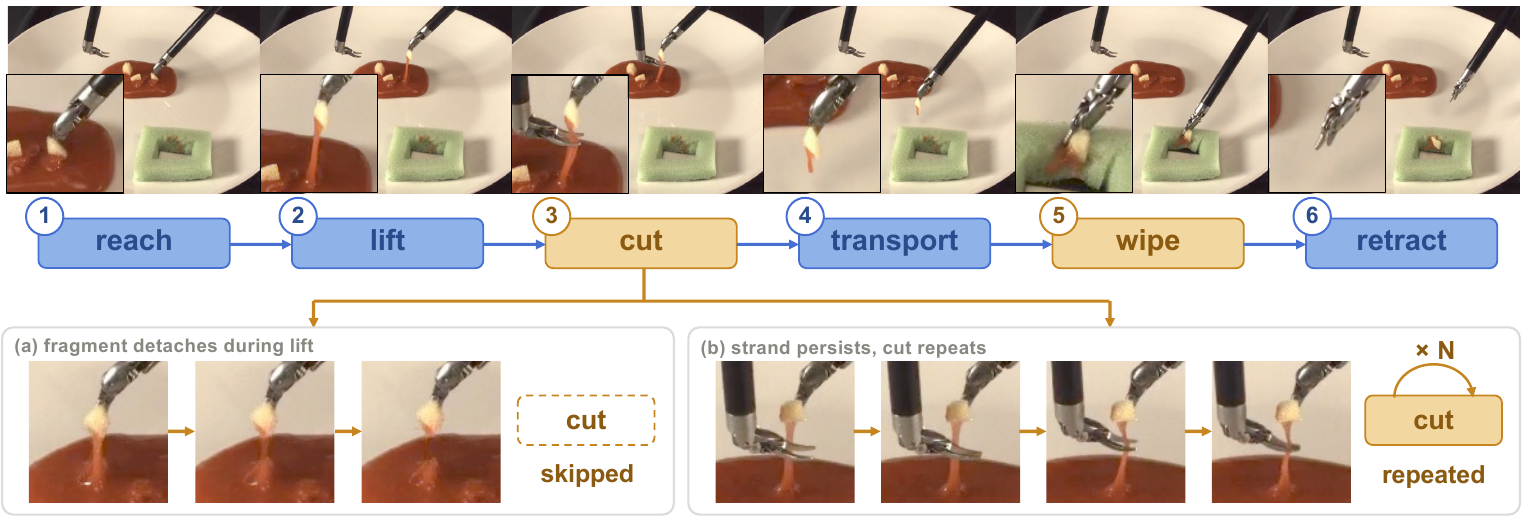}
    \caption{\small Illustration of one debridement cycle and the six identified skills, shown for the acting arm only; throughout the cycle, the other arm executes ``stabilize", holding its pose. 
    Model-based skills are shown in blue and policy-based skills in orange. The lower panels show two cases handled by the monitoring module at runtime: 
    (a) The fragment detaches during the ``lift", so the ``cut" skill is skipped; (b) The strand remains attached, so the ``cut" skill is activated repeatedly until it is severed.
    }
    \label{fig:experiment}
\end{figure*}
\paragraph{Model-Free Policy-Based Skills}
Every skill is first implemented as a procedural model-based skill and executed three times on the physical robot under human supervision, both to confirm safe operation and to measure reliability. 
A skill succeeding in fewer than two of three trials is re-implemented as a policy-based skill. 
Training data for bimanual tasks is typically collected in an unconstrained manner: the operator teleoperates both arms continuously, with no explicit boundary between phases of the task (\textit{Fig.}~\ref{fig:dataA}). 
To ensure that model-free policy-based skills chain with model-based ones without adjustment, we propose a compositional protocol that bounds every collected segment by the same identified conditions, as shown in \textit{Fig.}~\ref{fig:dataB}.
The agent drives the arms to a state satisfying \texttt{PRE} before recording begins, hands control to the human teleoperator with \texttt{DESC} as the instruction for the skill, and resumes control once the operator releases to verify that \texttt{POST} holds. 
The recorded entry poses are not identical but spread across the set \texttt{PARAMS} admits. 
Since deterministic policies such as~\cite{zhao2023learning} reproduce the same trajectory when reissued from the same state, a retry is unlikely to differ from the original attempt unless the entry pose does.
Under this protocol the operator teleoperates only the arm the skill acts on, and only the segments needed for that skill are recorded.

\begin{table*}[!t]
\centering
\caption{\small Comparison study on 20 batches of three-fragment removals (60 fragments in total). 
Bold indicates best performance.}
\label{tab:experiment}
\setlength{\tabcolsep}{5pt}
\renewcommand{\arraystretch}{1.15}
\begin{tabular}{l c c c c c c c c}
\toprule
Method &
\begin{tabular}[b]{@{}c@{}}Single-\\Fragment\\Success Rate\end{tabular} &
\begin{tabular}[b]{@{}c@{}}Consecutive\\Three-Fragment\\Success Rate\end{tabular} &
\begin{tabular}[b]{@{}c@{}}Three-Fragment\\Success Rate with\\One Human Intervention\end{tabular} &
\begin{tabular}[b]{@{}c@{}}Mean Time\\per Fragment (s)\end{tabular} &
\begin{tabular}[b]{@{}c@{}}Throughput\\(fragments/h)\end{tabular} &
\begin{tabular}[b]{@{}c@{}}Reaching\\Failure\end{tabular} &
\begin{tabular}[b]{@{}c@{}}Cutting\\Failure\end{tabular} &
\begin{tabular}[b]{@{}c@{}}Wiping\\Failure\end{tabular} \\
\midrule
Model-Based & 42\% & 15\% & 30\% & \textbf{26} & 58 & 14 & 18 & 3 \\
Model-Free        & 18\% & 0\% & 10\% & 77 & 8& 35 & 14 & 0 \\
\midrule
AGRO-SUVIDE & \textbf{85\%} & \textbf{60\%} & \textbf{95\%} & 47 & \textbf{65} & 2 & 5 & 2 \\
\bottomrule
\end{tabular}
\begin{tablenotes}
\footnotesize
\item Failures are conditionally dependent: cutting is attempted only after a successful reach, and wiping only after both. Counts therefore indicate where each method breaks down, not the relative difficulty of each stage.
\end{tablenotes}
\vspace{-1em}
\end{table*}

\subsection{Monitoring Module}

At runtime, the monitoring module pairs the skills $\mathcal{K}$ into phases $\Phi$, and composes the loop-style graph $\mathcal{L}$ sized to the $n$ fragments it observe. 
It then executes $\mathcal{L}$, verifying \texttt{PRE} and \texttt{POST} of each skill within a phase. 
Execution advances when \texttt{POST} is satisfied; when it is not, the monitor samples a new pose satisfying \texttt{PRE} and retries the skill.

\section{EXPERIMENTS}

\subsection{Implementation Details}

\subsubsection{Experimental Setup}
The full task sequence is illustrated in Fig.~\ref{fig:experiment}. 
Demonstrations are collected by teleoperating the PSMs from the surgeon console using two Master Tool Manipulators (MTMs) and a foot pedal set. 
An Allied Vision stereo RGB camera captures $1280 \times 960$ images at 30 fps. 
The task environment is a viscoelastic gel substrate mixed from clear PVA glue, borax activator, glycerin, and color pigment. 
Foam fragments placed in the substrate serve as debris targets, irregular in shape and ranging from 5 to 8 mm in maximum dimension.
In AGRO-SUVIDE, ``cut'' and ``wipe'' are model-free policy-based skills; the remaining skills are coded by the agent against $\mathcal{B}$.
Each model-free policy-based skill is trained with Action Chunking with Transformers (ACT)~\cite{zhao2023learning} on 33 three-fragment debridement episodes collected under the compositional data collection protocol. 
We compare AGRO-SUVIDE against two baselines. The first is an end-to-end bimanual policy trained with Action Chunking with Transformers (ACT)~\cite{zhao2023learning} (Model-Free), a widely used architecture for fine-grained bimanual manipulation~\cite{kim2024surgical,haworth2026suturebot}. 
The second is a procedural baseline~\cite{murali2015learning} (Model-Based), in which debridement is expressed as a hand-specified finite state machine over fixed motion skills.
All ACT-based policies trained in this paper takes the left image from the Allied Vision camera, resized to $224 \times 224 \times 3$, together with the PSM kinematic joint angles, and outputs absolute joint angles commanding both arms. 
We evaluate primarily on three-fragment debridement scenario ($n=3$) and include an additional five-fragment debridement scenario ($n=5$) as a held-out generalization test.

\subsubsection{Performance Metrics}

(a) \textit{Single-Fragment Success Rate}: the proportion of attempted fragments that complete the full debridement cycle. 

(b) \textit{Three-Fragment Success Rate}: the proportion of batches in which all three fragments are placed in the receptacle sequentially without failure. 

(c) \textit{Three-Fragment Success Rate with One Human Intervention}: as in (b), except that each trial carries a budget of one operator action. A reaching failure allows the arm to re-home and reattempt the grasp; a cutting failure allows the operator to sever the strand with surgical scissors manually; a wiping failure allows the operator to clear the adhered fragment from the jaws. Execution then resumes autonomously, and the budget does not refresh.

(d) \textit{Mean Time per Fragment (s)}: the average duration of a successful removal, from the arms leaving their home poses to the fragment being released into the receptacle.

(e) \textit{Throughput (fragments/h)}: The number of fragments successfully removed per hour.

We additionally record how often each of the three failure modes occurs:

(f) \textit{Reaching Failure}: the fragment is never lifted clear of the substrate, whether the grasp misses or slips. 

(g) \textit{Cutting Failure}: the raised strand survives the scissors, either missed or left partially attached. 

(h) \textit{Wiping Failure}: adhesive gel retains the fragment in the jaws and it never reaches the receptacle.

\subsection{Comparison Study} 
\textit{Tab.}~\ref{tab:experiment} compares AGRO-SUVIDE against the procedural model-based baseline and the model-free bimanual policy baseline. 
AGRO-SUVIDE reaches an 85\% single-fragment success rate and 60\% on consecutive three-fragment removal, rising to 95\% when one human intervention is permitted. 
Most remaining failures are incomplete cuts. 
The residual strand is a thin transparent film the monitor does not detect, so \texttt{POST} passes and the fragment is transported while still being partially attached.
Cutting also dominates the model-based baseline's failures. 
Following the original formulation, the scissors are commanded to a fixed offset 2.5 cm below the gripper in a shared frame, and since cable-driven dVRK arms exhibit pose-dependent kinematic error~\cite{kim2024surgical}, the commanded pose is often laterally offset from the strand, so the cut misses or leaves it partially attached. 
Reaching failures arise from a second source: specular glare on the gel is occasionally segmented as a fragment, and with no runtime check the pipeline commits to the false target and continues through the remaining phases. 
The model-free baseline fails earliest and most often, with 35 of its failures at reaching. When execution deviates from the demonstrated distribution, the policy does not recover, and typically hovers or drifts away from the substrate rather than advancing. 
We attribute this to the task horizon combined with scene variability: the gel is re-cast between batches, so substrate appearance and fragment placement differ every trial.
AGRO-SUVIDE is slower per fragment than the model-based baseline at 47 seconds versus 26 seconds, since retried skills incur additional policy rollouts, but its higher success rate compensates. 
The model-free baseline is slowest on both counts, combining a low success rate with per-step inference overhead. 

\subsection{Ablation Study}
We additionally ablate two components of AGRO-SUVIDE.

\begin{figure}[t]
    \centering
  \begin{subfigure}[h]{0.175\textwidth}
    \centering
    \includegraphics[width=\textwidth]{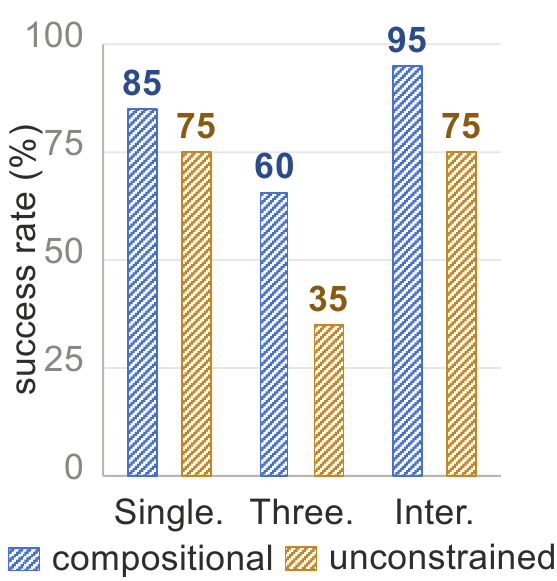}
    \vspace{-1.8 em}
    \caption{}
    \label{fig:ablation_data}
  \end{subfigure}
  \begin{subfigure}[h]{0.3\textwidth}
    \centering
    \includegraphics[width=\textwidth]{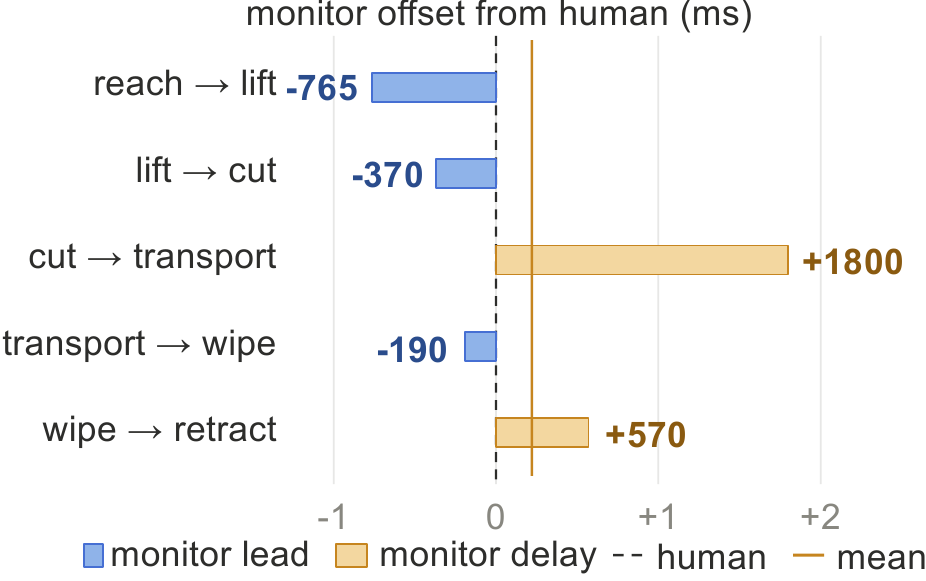} 
      \vspace{-1.8em}
    \caption{}
    \label{fig:ablation_monitor}
  \end{subfigure}
    \caption{\small Ablation study. (a) Success rates with and without the compositional collection protocol. (b) Monitor transition timing against a human observer.}
    \label{fig:ablation_monitor2}
    \vspace{-1em}
\end{figure}

\paragraph{Compositional Data Collection Protocol}
We retrain the ``cut" and ``wipe" policies on 33 episodes each collected without condition boundaries, and evaluate the resulting system against the original over 20 batches of three-fragment debridement (60 fragments).
Segments begin and end wherever the operator chooses, guided only by \texttt{DESC} to ensure the skill content is covered. 
Success rates fall across all three metrics relative to policies trained under the compositional protocol (\textit{Fig.}~\ref{fig:ablation_data}).
Three factors likely contribute to this drop.
First, removing the boundary widens the state distribution the policy must cover, so the same 33 episodes may spread thinly over a larger space. 
Second, the resulting policies less reliably reach the pose specified by the post-condition. 
After a successful cut, PSM2 sometimes remains in place rather than withdrawing, so the strand is severed but the arm never clears the workspace, and the transport phase would advance PSM1 into a colliding pose, prompting the human supervisor to stop execution. 
Third, the monitor cannot confirm a post-condition it never observes, so it waits through the timeout and retries, terminating after five attempts.
\paragraph{Monitor}
We additionally compare the monitor's transition decisions against a human observer over 10 batches of three-fragment debridement (30 fragment removals). 
To isolate the comparison from sensing rate, the observer sees the robot only through the same camera stream the monitor receives, and presses a key at each transition.
The monitor's mean absolute deviation from the human is 221 ms; per-transition leads and lags appear in
\textit{Fig.}~\ref{fig:ablation_monitor}. 
The monitor is faster than the human on model-based skill transitions, where the terminating pose is commanded rather than produced, so the conditions are met the moment the pose is reached.
The monitor lags only on the two transitions following model-free skills, where the terminating state is produced by the policy, so the transition waits until \texttt{POST} is reached.

\subsection{Generalization Study: Five-Fragment Scenarios} 
\begin{figure}[h]
    \centering
    \includegraphics[width=\linewidth]{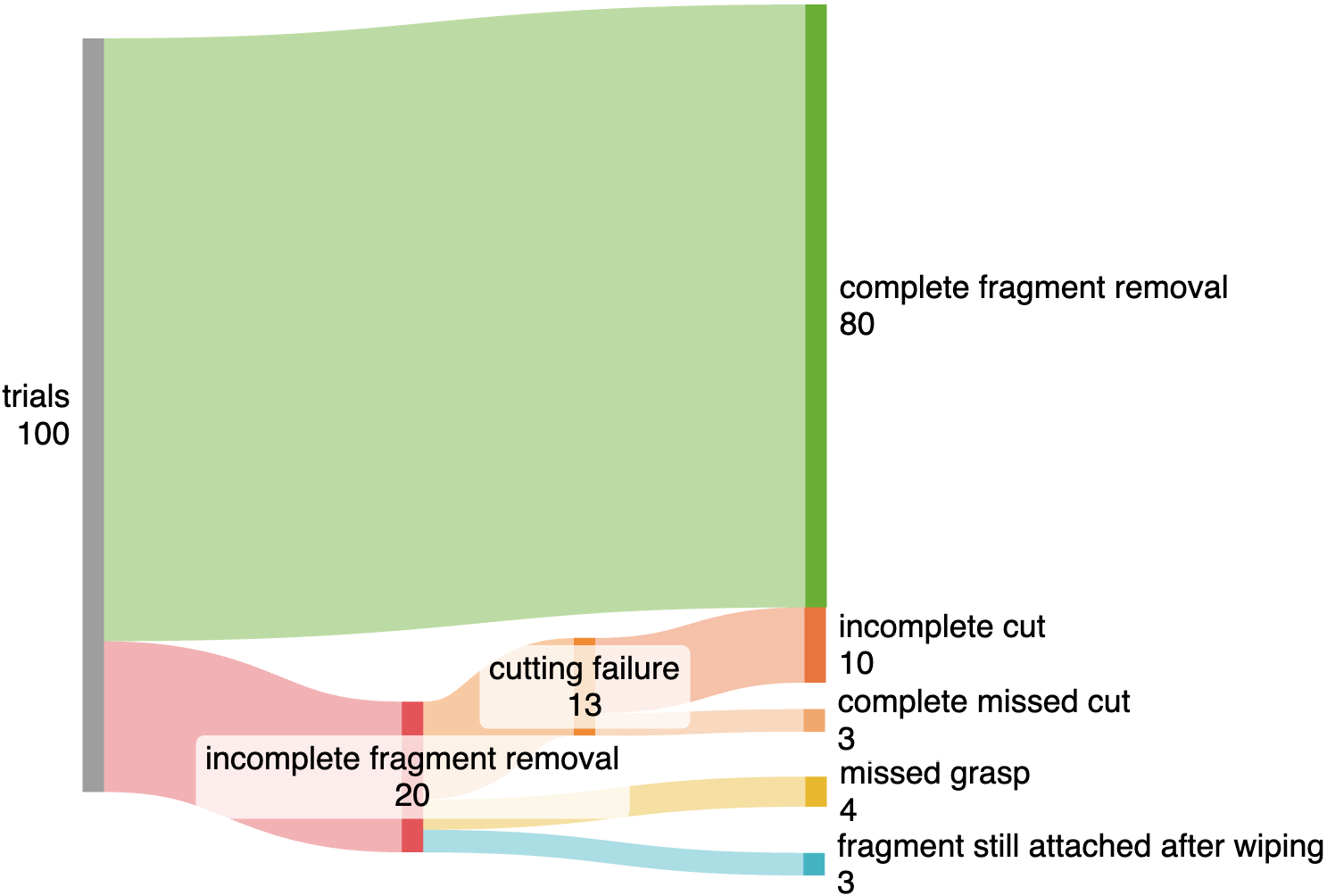}
    \caption{\small A Sankey diagram of the five-fragment generalization test over 20 batches (100 fragments) by AGRO-SUVIDE.}
    \label{fig:generalisation}
    \vspace{-1em}
\end{figure}

We test AGRO-SUVIDE on five-fragment debridement over 100 fragment removals ($n=5$), with results and failure attributions shown in \textit{Fig.}~\ref{fig:generalisation}. 
Neither component of the system has encountered this condition: the conditions are identified from a three-fragment demonstration ($N=3$), and the model-free policy-based skills are trained only on three-fragment episodes. 
The single-fragment success rate in unseen five-fragment scenarios is 80\%.

\section{LIMITATIONS AND FUTURE WORK}
AGRO-SUVIDE identifies its skills from a single three-fragment removal demonstration, which we assume to be expert and procedurally repetitive. 
Procedures whose steps vary in order between repetitions, or that contain intra-procedure recovery, fall outside the current scope. 
The reliance on repetition is consistent with clinical convention, since surgical subtasks are typically performed in a standardized sequence. 
Future work will extend the skill identification and execution loop to other repetitive surgical tasks, and enrich the loop-style graph with nodes that capture the recovery behaviors present in human demonstrations.

\section*{ACKNOWLEDGMENT}
This work was partially supported by the Wallenberg AI, Autonomous Systems and Software Program (WASP) funded by the Knut and Alice Wallenberg Foundation.

\bibliographystyle{IEEEtran}
\balance
\bibliography{citations}
\end{document}